\pdfoutput=1
\documentclass[twocolumn]{cinc}
\usepackage{graphicx}
\usepackage{textcomp}
\begin{document}
\bibliographystyle{cinc}

\title{Early Detection of Sepsis using Ensemblers}

\author {
Shailesh Nirgudkar \textsuperscript{1}, Tianyu Ding  \textsuperscript{1, 2}\\
\ \\
\textsuperscript{1} MathWorks, Natick, MA, USA\\
\textsuperscript{2} Johns Hopkins University, Baltimore, MD, USA}

\maketitle

\begin{abstract}
This paper describes a methodology to detect sepsis ahead of time by analyzing hourly patient records. The Physionet 2019 challenge consists of medical records of over 40,000 
patients. Using imputation and weak ensembler technique to analyze these medical records and 3-fold validation, a model is created and validated internally.  
The model achieved an accuracy of 93.45% and a utility score of 0.271. The utility score as defined by the organizers takes into account true positives, negatives and false alarms.
\end{abstract}

\section{Introduction}
Sepsis is a life threatening condition in which a person's immune response to infection can cause tissue damage, multiple organ failure, and even death \cite{Singer}. If this 
condition is detected early enough, preventative steps can be taken to avoid a mortal outcome \cite{Kumar, Seymour}. Clinicians have revised the definition of sepsis in the hope 
that with it, relevant measurements can be recorded and an early diagnosis can be made. Still, the detection remains challenging because it is not possible to measure when the
'sepsis' starts. The onset can be observed only indirectly through vital signs, laboratory measurements, administration of antibiotics, and drawing of blood cultures for suspicion
of infection. In emergency departments, there are established protocols for regular measurements; however, in normal hospital settings such regularity is not observed. Typically, the
measurements are taken at irregular intervals and have missing information. Moreover, timely measurments are taken only in case of suspicion of disease. Prior art \cite{Henry} 
has been able to predict sepsis ahead of time using machine learning techniques but with good, clean data. This database contains patient records only from one hospital between 2001 and 2007 and some of the features in this dataset 
were defined by ICD-9 (International Classification of Diseases, Ninth revision) codes, the sensitivity and specificity of these codes were very diagnosis-dependent. The 
techniques may not scale without further work. In real life, data available for analysis are often noisy because of the reasons mentioned earlier. The labeled dataset provided by 
Physionet 2019 challenge \cite{Reyna} has more than 90\% values missing the laboratory measurements. Also the data are highly imbalanced meaning there are very few patients 
($\approx$7.3\%) having sepsis condition. The combination of high percentage of missing data and imbalance in class labels makes the problem of prediction challenging. In this paper, 
we describe an approach to preprocess the data and train ensemble learner on that data. 

\section{Methodology}
Since the training data contained significant missing values, the data are preprocessed. Out of 40 features per patient, features which have missing values are identified 
and the missing values are computed using imputation. 

\subsection{Feature Imputation}
\begin{figure}[h]
\centering
\includegraphics[width=6.8cm]{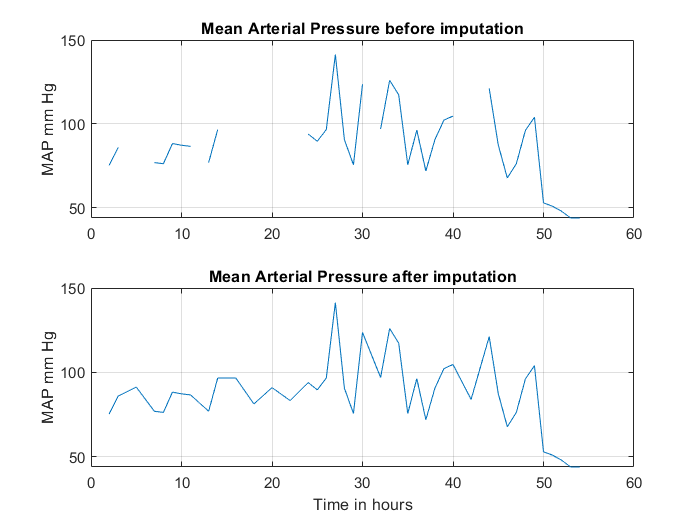}
\caption{Imputation of MAP of a patient record}
\label{fig:data_imputation}
\end{figure}

Each patient record consists of 40 features and a final label for each hour. Majority of features related to laboratory measurement contain over 90\% NaN values. These missing values 
are not missing at random (MNAR) but because the measurements are generally taken twice a day. Data imputation is used to generate missing values. For 
each patient, values are imputed using linear imputation within that patient record. As an illustration, imputation of mean arterial pressure (MAP) for a patient record is shown 
in Figure~\ref{fig:data_imputation}. If the number of non-NaN values are less than 3, that feature record is treated as noisy and no imputation is attempted. At the end of 
this phase, there would still be some features containing NaN values. The simple imputation (of linearization within a patient record) is chosen over other complicated schemes 
because there is no satisfactory method when data are not missing at random (MNAR) and there are many features exhibiting the issue.

\subsection{Feature Weights}
As per \cite{Singer}, mean arterial pressure (MAP) and serum lactate level are important in identifying septic shock. This combination alone is associated with over 40\% mortality
rates in hospital. At least two of the respiratory rate, altered mentation and systolic blood pressure (SBP) determine quickSOFA (:Sequential [Sepsis-related] Organ Failure 
Assessment (SOFA)) score. quickSOFA is a practical way of identifying adult patients in out-of-hospital, emergency departments or general hospital wards who are likely to develop
poor outcomes. So more weightage is given to features `Lactate', `MAP', `HR' and `Resp' by squaring them and replacing the original values with squared ones.

\subsection{Model}
Since the final training data set still contain missing values, surrogate trees are used, which can handle these values. MATLAB{\textsuperscript\textregistered}  release R2019a 
is used to implement the algorithm. The software sends the observation to left or right child node using the best surrogate predictor. Because of the imbalance in labels (sepsis vs non-sepsis) of training data, 
a variation of Boosting algorithm known as {\it Random UnderSampling Boost} ({\bf RUSBoost}) \cite{Seiffert} is used. It combines data sampling with AdaBoost algorithm. As the name 
implies, RUSBoost randomly undersamples the training data corresponding to the majority class. The main disadvantage of random undersampling, loss of information, is greatly overcome 
by combining it with boosting. RUSBoost automatically creates datasets where 35\%, 50\%, or 65\% of the examples in the post-sampling dataset are minority class examples. The 
algorithm then report the parameters that result in the best performance. This algorithm is simple and takes less training time and consumes less memory than than its variation 
SMOTEBoost \cite{Chawla}. Also the experimentation has shown us that for larger number of learning cycles the error loss is less but the model overfits the training data and gives
poor result on test data. Hence, to prevent the model from over-fitting, the number of learning cycles is limited to 200.

\section{Experimental results}
\begin{figure}[ht]
%\centering
\includegraphics[width=6.8cm]{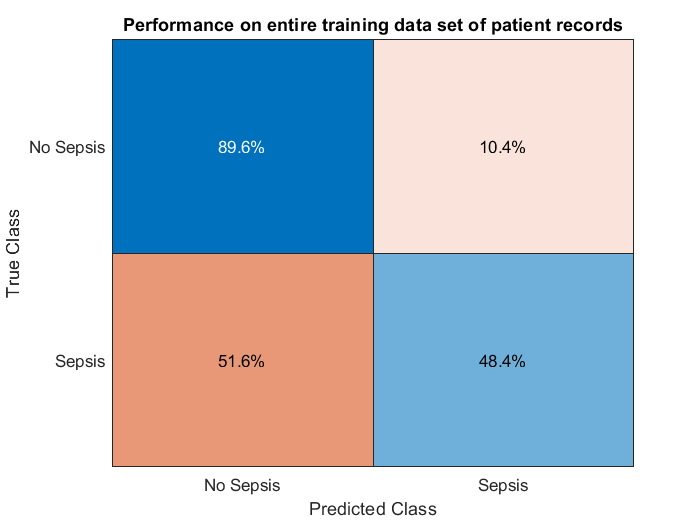}
\caption{Confusion chart for training data}
\label{fig:confusionchart}
\end{figure}

All the experimentation is performed on the Physionet 2019 challenge dataset \cite{Reyna}. The dataset consists of over 45,000 patient records. Each patient record 
consists of hourly reading of vital signs, lab results and other demographics totalling 40 features. 5,000 records were provided initially and later on two sets of 
approximately 20,000 records were made available. All the three datasets were combined into a single one and then it was divided into a training dataset (80\%) and 
an internal test dataset (20\%). We subdivided the training dataset into two parts: 1/3 is treated as cross-validation dataset and the remaining 2/3 is used as 
training dataset. The 3-fold validation scheme is chosen for efficiency. After model hyper-parameters are determined, the entire training dataset (36,000 records) 
is used to obtain the final model alongwith the utility score. Figure~\ref{fig:confusionchart} 
shows the performance of the model on the entire training dataset. The model metrics on internal test data set (9,000 records) is as shown in 
Table~\ref{table:internaldatautilityscore}. The utility score is a specific metric devised by the challenge committee to detect usefulness of a given algorithm in predicting sepsis 
ahead of time. The metric rewards true early detection (true positives) and punishes false alarms (false positives) as well as failing to detect disease when it is present 
(false negatives). 

\begin{table}[htbp]

\caption{\label{tab:font} Performance score on internal test data set}
\label{table:internaldatautilityscore}
\vspace{4 mm}
\centerline{\begin{tabular}{ccccc} \hline\hline
AUROC & AUPRC & Accuracy & F-measure & Utility \\ \hline
0.5282 & 0.0582 & 0.9337 & 0.1533 & 0.271 \\ \hline
\end{tabular}}

\end{table}

\begin{figure}[ht]
%\centering
\includegraphics[width=6.8cm]{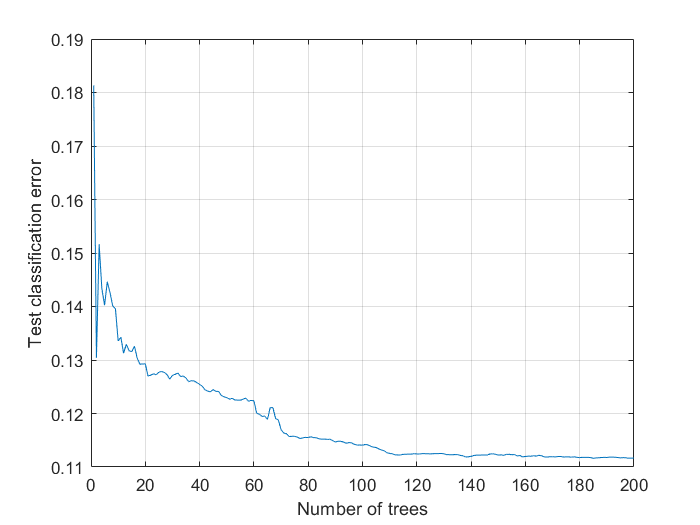}
\caption{Classification loss for training data}
\label{fig:loss}
\end{figure}

The score could not be improved further because significant amount of data are missing. Also, the representation of classes is highly imbalanced. Only about 7\% of patients are 
labeled as sepsis patients. We also observe that increasing number of weak learners or optimizing the hyper-parameters further overfits the training data and gives reduced results 
on hidden test data. Hence we limit the number of learning cycles to 200. We could get classification loss minimized as shown in Figure~\ref{fig:loss}.

This model is used to obtain utility score on the hidden dataset maintained by the Physionet organizers. On a hidden dataset maintained by organizers, a utility score of 0.192 was obtained.
Other statistics on a test data maintained by Physionet is as follows:

\begin{table}
\caption{Performance score on individual test data set}
\label{table:individualdatautilityscore}
\vspace{4 mm}
\centerline{\begin{tabular}{cccccc} \hline\hline
Test set & AUROC & AUPRC & Accuracy & F-measure & Utility \\ \hline
Set A & 0.579 & 0.057 & 0.929 & 0.171 & 0.274\\ \hline
Set B & 0.621 & 0.038 & 0.928 & 0.120 & 0.233\\ \hline
Set C & 0.626 & 0.014 & 0.765 & 0.036 & -0.246\\ \hline
\end{tabular}}
\end{table}

\section{Alternatives considered for improving score}
Various approaches are tried to improve the utility score.
\begin{itemize}
\item Drop features which have high percentage of NaN values: This resulted in lower utility score. This is because there is loss of information which affects
training of the model.
\item Removal of outliers: From entire dataset, based on inter quartile range of a feature, outliers are removed. It resulted in lower utility score. The 
authors did not investigate the root cause for the issue.
\item Feature creation: New features are created based on difference between hourly records of a feature. Since there is no new information this strategy 
did not result in improving the utility score.
\end{itemize}

\section{Conclusion}
We propose a methodology to detect onset of sepsis ahead of time. Model applied on the Physionet 2019 challenge dataset obtained a utility score of 0.274. The code is available as open source software under GNU license. We plan to refine the imputation strategy \cite{Camino} and also employ deep learning tools \cite{Lauritsen} which will help in improving classification accuracy. On a related but different note, if nursing assessements are available Rothman Index \cite{Rothman} can be computed which is a better predictor in sepsis detection \cite{Rothman2}.

\section{Conflict of interest statement}
Shailesh Nirgudkar is employed at MathWorks and Tianyu Ding was an intern at MathWorks at the time of this work.

\bibliography{refs}

\vspace*{4mm}

\vspace*{-.6cm} %adjustment to make the columns even
\begin{correspondence}
Shailesh Nirgudkar\\
Control Design Automation, MathWorks,\\
1 Lakeside Campus Drive, Natick, MA 01760 USA.\\
shailesh.nirgudkar@gmail.com
\end{correspondence}
\end{document}